\documentclass[12pt]{article}

\usepackage{multirow}
\usepackage{graphicx}
\usepackage{caption}
\usepackage{times}
\usepackage{amsmath}
\usepackage[colorlinks=true, linkcolor=blue, citecolor=blue, urlcolor=blue]{hyperref}
\usepackage{makecell}

\pdfoutput=1

\begin{document} 

\title{Cross-Domain Foundation Model Adaptation: 
Pioneering Computer Vision Models for Geophysical Data Analysis}


\author
{Zhixiang Guo$^{1}$, Xinming Wu$^{1\ast}$, Luming Liang$^{2}$, \\
Hanlin Sheng$^{1}$, Nuo Chen$^{1}$ and Zhengfa Bi$^{3}$\\
\normalsize{$^{1}$School of Earth and Space Sciences, 
University of Science and Technology of China,}\\
\normalsize{Hefei, 230026, China}\\
\normalsize{$^{2}$Microsoft Applied Sciences Group, 
Redmond, WA 98052, United States}\\
\normalsize{$^{3}$Lawrence Berkeley National Laboratory, 1 Cyclotron Rd, CA 94707, USA}\\
\\
\normalsize{$^\ast$To whom correspondence should be addressed:}\\
\normalsize{E-mail: xinmwu@ustc.edu.cn}
}
\date{}






\maketitle

\begin{abstract}
We explore adapting foundation models (FMs) from the computer vision domain to geoscience.
FMs, large neural networks trained on massive datasets, excel 
in diverse tasks with remarkable adaptability and generality. 
However, geoscience faces challenges like lacking curated 
training datasets and high computational costs for developing 
specialized FMs. This study considers adapting FMs from computer 
vision to geoscience, analyzing their scale, adaptability, and 
generality for geoscientific data analysis. We introduce a 
workflow that leverages existing computer vision FMs, fine-tuning 
them for geoscientific tasks, reducing development costs while 
enhancing accuracy. Through experiments, we demonstrate this 
workflow's effectiveness in broad applications to process and 
interpret geoscientific data of lunar images, seismic data, 
DAS arrays and so on. Our findings introduce advanced ML 
techniques to geoscience, proving the feasibility and 
advantages of cross-domain FMs adaptation, driving 
further advancements in geoscientific data analysis 
and offering valuable insights for FMs applications 
in other scientific domains.
\end{abstract}

\section*{Introduction}
Foundation models (FMs) refer to deep learning models with millions to billions 
of parameters, pre-trained on massive datasets containing 
tens of millions to billions of data~\cite{bommasani2021opportunities}.
Training FMs on large, diverse datasets that cover a wide range 
of scenarios enables them to develop comprehensive and 
adaptable representations. 
This leads to state-of-the-art (SoTA) results and exhibits significant 
generalization capabilities for few-shot and zero-shot 
tasks~\cite{oquab2023dinov2,kirillov2023segment}.
Therefore, FMs usually serve as base models to develop models for 
various task types efficiently and effectively, demonstrating  
significantly superior performance and generalization 
across tasks and datasets compared to traditional ML algorithms 
trained on specific datasets for specific tasks~\cite{aws}. 
In recent years, foundation models have made significant advancements 
in fields such as natural language processing~\cite{devlin2018bert,
brown2020language,touvron2023llama}, 
computer vision~\cite{yuan2021florence,bommasani2021opportunities,li2024multimodal}, 
healthcare~\cite{zhou2023foundation,moor2023foundation,huang2023visual,thieme2023foundation},
autonomous driving~\cite{cui2024survey,chen2023end} 
and so on, distinguishing themselves from 
traditional ML algorithms with their remarkable adaptability 
and generalizability~\cite{bommasani2021opportunities}.  
The research on foundation 
models has revolutionized the development of artificial intelligence (AI), 
representing a crucial trend for the future of 
AI~\cite{schneider2024foundation}.

Deep learning methods have found extensive applications
in the field of geophysics, including 
seismology~\cite{ross2018p,zhu2019phasenet,pardo2019seismic,mousavi2020earthquake,mousavi2022deep,si2024all}, 
earthquake monitoring~\cite{perol2018convolutional,mousavi2019bayesian,rouet2020probing,zhu2022earthquake,yang2022toward,wang2022predicting},
earthquake forecasting~\cite{johnson2021laboratory,beroza2021machine,laurenti2022deep,convertito2024deep}, 
seismic data processing~\cite{yuan2018seismic,yu2019deep,ovcharenko2019deep,wang2019deep,chai2020deep,park2020automatic,mousavi2024applications}, 
interpretation~\cite{di2018multi,qian2018unsupervised,wrona2018seismic,wu2019faultseg3d,pham2019automatic,tolstaya2022deep,wu2023sensing}, 
and inversion~\cite{yang2019deep,li2019deep,liu2021deep,zhang2023domain} and more.
However, these methods generally adopt the development 
of task-specific deep learning models, 
facing challenges in generalization across 
different tasks and even different regions~\cite{lu2019deep,yu2021deep}. 

Foundational models (FMs), known for their 
generality and versatility, offer a promising solution 
to these challenges.
Yet, research on FMs in geophysics is 
limited~\cite{sheng2023seismic}, 
facing significant challenges, particularly in the 
construction of large-scale datasets, 
the immense computational resources required, 
and the high associated energy costs for training these models.
Firstly, constructing large-scale, well-curated, 
and comprehensive training datasets is a major obstacle. 
As noted by Myers et al. (2024), the success of FMs in other 
domains largely relies on the availability of extensive public datasets.
Popular visual FMs, such as CLIP (400M images)~\cite{radford2021learning},  
MAE (1.3M images)~\cite{he2022masked}, 
SAM (11M images)~\cite{kirillov2023segment}, 
and DINOv2 (142M images)~\cite{oquab2023dinov2}, 
rely on vast datasets up to hundreds millions of training samples. 
In geophysics, however, the confidentiality of data, 
often involving sensitive information related 
to resource exploration and regional topography, 
presents a significant barrier to public dissemination~\cite{tenopir2018research}. 
The economic value of these data further complicates its disclosure. 
The low public availability of geophysical datasets makes it extremely 
challenging to collect large-scale datasets. 
Additionally, the diversity in geophysical data acquisition systems, 
non-standardized and uncertain data processing workflows, variations 
in noise and geological backgrounds, data sampling intervals, 
data value distributions, and frequency band distributions, 
all pose substantial difficulties for data cleaning and curation, 
making it hard to form a standardized, comprehensive dataset. 
Secondly, the computational resources and time costs 
required to train FMs are prohibitively high.
Training FMs, which involve hundreds of millions 
of parameters, typically demands hundreds to thousands 
of GPUs and several months of processing time~\cite{jiang2024megascale}. 
This substantial investment creates high entry barriers, 
limiting the capability to a few financially robust companies. 
In geophysics, even fewer companies possess the necessary 
resources, and the uncertain return on investment further deters such endeavors.
Consequently, the 
development of FMs in the geophysics field remains 
relatively undeveloped at present.

\begin{figure*}[t]
\centering
\includegraphics[width=\textwidth]{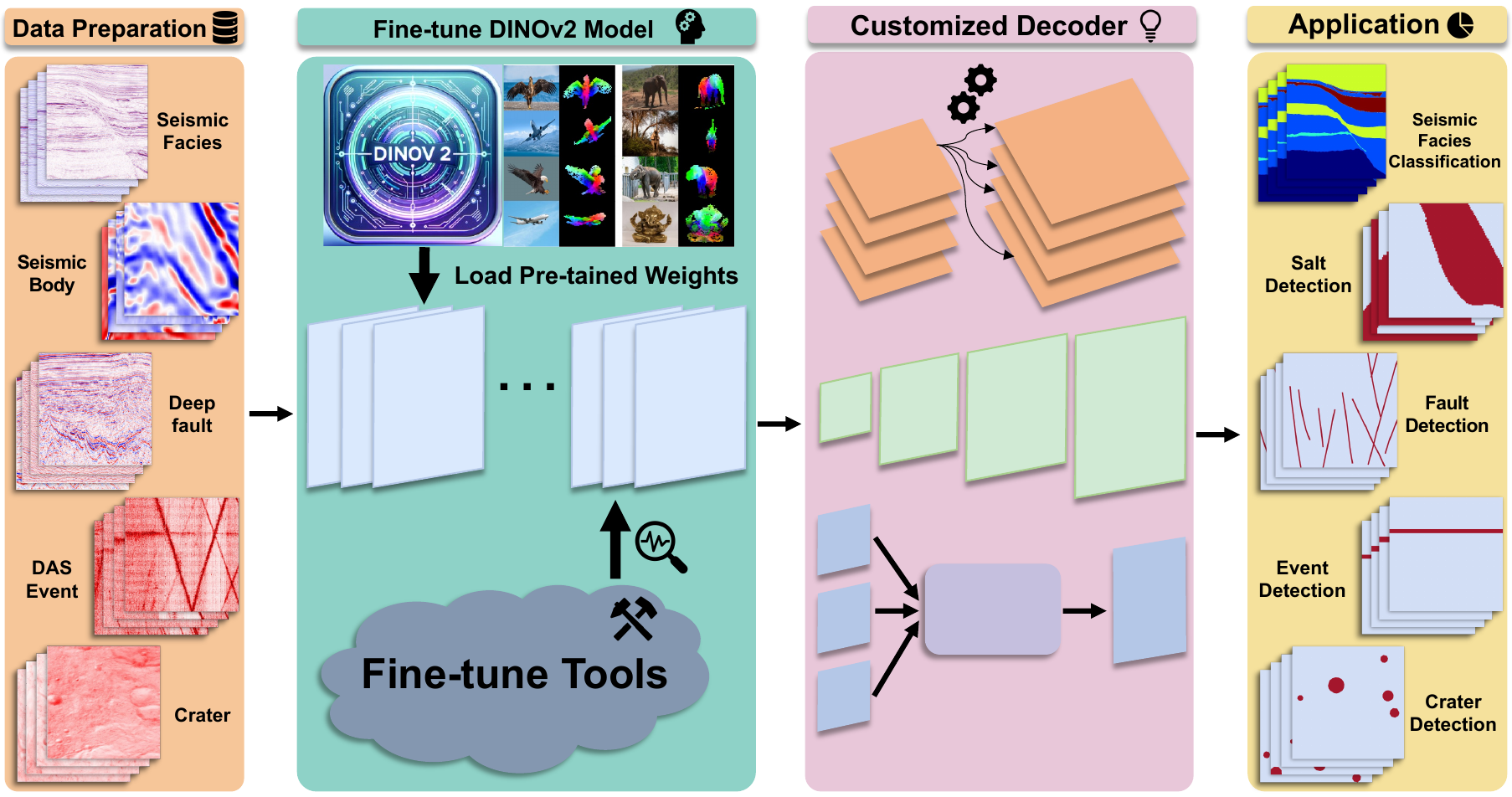}
\caption{Workflow for adapting pre-trained foundation models to geophysics.
First, we prepare geophysical training datasets (1st column), 
which involves collecting and processing relevant geophysical data 
to ensure it is suitable for adaption fine-tuning. Next, we load the pre-trained 
foundation model as the data feature encoder (2nd column) 
and fine-tune the model to make it adaptable to geophysical data. 
To map the encoder features to the task-specific targets, 
we explore suitable decoders 
(3rd column) for geophysical downstream adaption. Finally, the adapted model 
is applied to various downstream tasks within the geophysics 
field (4th column).}
\label{fig:workflow}
\end{figure*}

In light of the challenges in training FMs for geophysics and the 
similarity between geophysical data and natural images, we 
propose adapting mature visual FMs to this domain.
This approach aims to reduce the dataset requirements 
and lower computational and time costs for developing 
FM applications in geophysics.
We presented a comprehensive FMs adaptation workflow 
(Fig.~\ref{fig:workflow}) including data preparation, 
fine-tuning the foundation model, and selecting suitable decoders 
for downstream tasks.
First, 
we collected several types of typical geophysical 
data that vary significantly in data types, quantities, and sizes 
to explore how these aspects affect the foundation model adaptation. 
During the adaptation process, we found that the foundation model could be flexibly applied 
to various data types.
Remarkably, only a small number of samples were needed to adapt the 
large models. This feature is particularly advantageous for geophysical 
applications, where many scenarios often have a limited number of labeled samples.
Second, 
we explored the pre-trained vision foundation model's capability 
for feature extraction and representation of geophysical data, 
and further enhanced this capability by fine-tuning the model 
using a parameter-efficient method.
This method enhanced the model's understanding 
of geophysical data effectively and efficiently with minimal parameter updating.
Finally, we fed the high-dimensional features of geophysical data, 
represented by the foundation model, into several decoders with different structures. 
These decoders recover the target information embedded 
in the high-dimensional features to produce the results 
required for downstream tasks. By analyzing our test results, 
we provided recommendations and references for using different 
decoders for different downstream tasks.

\section{Results}\label{sec2}
\subsection{Choice of pre-trained FM}
The development of visual FMs closely 
follows the advancements in large language models, 
leveraging proxy tasks to pre-train large-parameter models for 
deep feature understanding of images. One of the earliest 
visual FMs, MAE~\cite{he2022masked}, learns rich hidden 
representation of natural images and visual concepts through 
pre-training by the self-supervised learning strategy of 
randomly masking patches of input images and reconstruct them.
Language text and images are not 
isolated, humans often summarize images in textual form, 
with this in mind, the text-image multimodal foundation 
model CLIP~\cite{radford2021learning} was developed. 
Training multimodal models is highly practical, but it 
requires a large amount of multimodal datasets for training, 
which poses a significant challenge.
Traditional ML algorithms typically produce fixed outputs once 
trained, which may result in outcomes that do not meet the user's 
expectations. To address this, 
SAM~\cite{kirillov2023segment} introduces human 
prompts into deep neural network inference, 
enabling real-time updates to the model's outputs 
and progressively achieving the desired results, 
but it requires using a large amount of prompt 
and label pairs to train and integrate the prompts 
into the decoder, making 
the training more challenging.
The ultimate goal of AI is for machines to understand 
the world like humans. DINOv2~\cite{oquab2023dinov2}, 
pre-trained by a discriminative self-supervised 
contrastive learning scheme, aims to understand the global 
features of images while also paying attention to local details.
DINOv2 has a profound understanding of images, 
outperforming other FMs in various benchmarks such as 
image semantic segmentation, image classification, 
depth estimation and so on~\cite{oquab2023dinov2},
especially in few-shot semantic segmentation~\cite{bensaid2024novel}, 
which is highly significant for applications in geophysics due to
the lack of labelled large datasets for fine-tuning.

While we could use any other visual foundational model as 
a base for developing geophysical downstream task applications, 
we have chosen DINOv2 for this paper due to its superior 
feature extraction and representation capabilities compared 
to other foundational models (Oquab et al., 2023).
To equip the pre-trained model with robust feature extraction 
and representation capabilities for natural images, 
DINOv2 primarily made efforts in the following aspects:
Firstly,
DINOv2 builds upon DINO~\cite{caron2021emerging} by 
integrating self-supervised pre-training techniques 
from both DINO and iBOT~\cite{zhou2021ibot}. 
DINO's self-supervised contrastive learning approach, which is based 
on image-level objectives, allows the network to effectively learn 
global features and reduce training fluctuations. On the other hand, 
iBOT's patch-level approach emphasizes the network's attention to 
local details. Moreover, DINOv2 used SwAV~\cite{caron2020unsupervised} normalization 
to stably integrate these two methods into training, thereby balancing 
both global and local features.
Secondly,
to enable the model to learn rich and non-redundant 
image features, DINOv2 has made significant efforts 
in data curation. These efforts include deduplication~\cite{pizzi2022self}, 
self-supervised image retrieval~\cite{johnson2019billion} to construct 
the dataset. This process ultimately yields a curated dataset of 
142 million nondundant and diverse images (LVD-142M) 
from the widely collected 1.2 billion images for more 
effectively and efficiently training.  Pretraining with this 
curated dataset, DINOv2 shows significantly better performance 
than DINO that was pre-trained with the originally 
uncurated large dataset of 1.2 billion images~\cite{oquab2023dinov2}.
Finally,  
based on the largest pre-trained model, 
multiple variants of DINOv2s are distilled for more efficient downstream applications. 
This reduces the model parameters while 
maintaining performance~\cite{hinton2015distilling}, making it more effective 
and efficient for applications in the field of geophysics.
The efforts in the above three aspects have enabled DINOv2 to surpass 
some weakly supervised and supervised learning methods 
in terms of generalization across datasets, as well as in few-shot and zero-shot tasks, 
using only simple decoders such as linear probing and kNN~\cite{oquab2023dinov2}.

In summary, DINOv2 possesses the ability to extract both global and 
local features, broadly generalize across datasets, and efficiently 
extract features by self-distilling into smaller models. Next, 
we will use DINOv2 as a base to explore how to effectively adapt 
a vision foundation model to downstream tasks in the 
geophysical domain.

\begin{figure*}[htp]
\centering
\includegraphics[width=.5\textwidth]{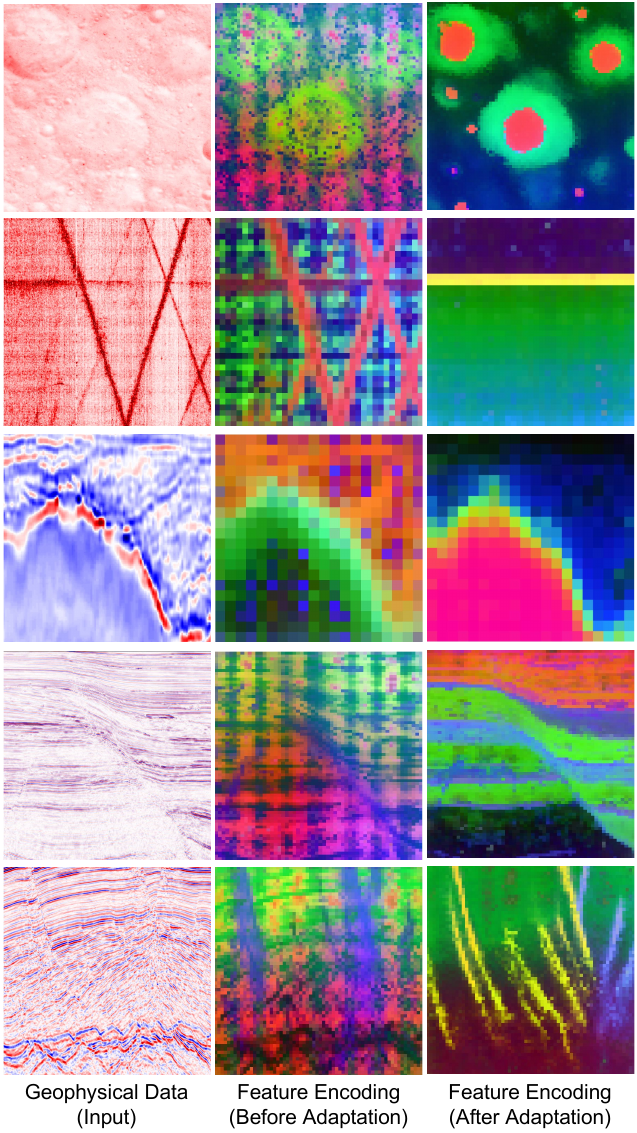}
\caption{DINOv2's feature representation of geophysical data. 
The 1st column shows typical geophysical data, 
including, from top to bottom, lunar images containing craters, 
DAS data with seismic events, seismic data with salt domes, 
strata facies, and deep 
faults. We input these data into the pre-trained 
DINOv2, which serves as an encoder to compute the feature 
representation of the data. The RGB visualization shows 
the three most representative components of the geophysical 
data feature representation by the pre-trained DINOv2 
before (2nd column) and after (3rd column) fine-tuning. 
We observe that DINOv2, initially pre-trained on natural 
images, exhibits a general capability for representing 
geophysical data features, forming a basis for its 
adaptation to geophysical tasks. Fine-tuning further 
enhances this feature representation (3rd column), ensuring 
advanced performance in geophysical applications.}
\label{fig:PCA}
\end{figure*}

\subsection{Pre-trained FM for geophysical data feature representation}
To further validate the potential of DINOv2's adaptation to the geophysics 
field, we directly used the pre-trained DINOv2 (ViT-S/14 with knowledge distillation) 
as an encoder to explore its capability of feature representation of geophysical data, 
including lunar images (containing craters), 
DAS data (containing seismic events), 
and seismic data (containing seismic facies, geobodies, and faults),
as shown in the first column of Fig.~\ref{fig:PCA}. 
To better understand and visualize the features computed by 
the DINOv2 for the hidden representation of the geophysical data, 
we performed principal component
analysis (PCA) on them.
PCA can reduce the dimensionality of the high-dimensional features 
output by the encoder, identifying the components with the 
most significant characteristics of the features. In this way, 
we reduce the high-dimensional features to three most significant 
components and visualize them as RGB colors in the second 
column of Fig.~\ref{fig:PCA} to
understand the latent space representation of geophysical 
data by DINOv2.
This feature representation in the latent space forms 
the basis and potential capability of the model for subsequent downstream tasks. The better the 
model expresses the data features, the better it can perform 
downstream tasks.

As shown in the second column of Fig.~\ref{fig:PCA}, DINOv2, 
despite being trained on only natural images, still shows 
general capability to extract and represent key features 
of previously unseen geophysical data. 
For examples, the most dominant targets (lunar caters, 
DAS events, and salt bodies in the first three images 
in the 2nd column of Fig.~\ref{fig:PCA}) within the 
geophysical data can be effectively expressed in the latent space. 
We believe that this is because the multidimensional 
features of geophysical data are somewhat similar to 
those of natural images.
This poses the basis and potential that DINOv2 
is adaptable for geophysical applications. 
However, the subtle targets like strata and faults 
are poorly represented as shown in the last three 
images of the 2nd column of Fig.~\ref{fig:PCA}.  
Moreover, the distinction between targets and background 
is not clear enough, and some noisy features or artifacts are apparent.
Based on these observations, DINOv2 demonstrates a basic 
capability to understand and represent geophysical data, 
but this capability is not perfect. DINOv2's ability to 
represent geophysical data is limited because 
its pre-training data primarily consists of natural 
images, lacking geophysical data. Consequently, 
it has not learned the key features of geophysical 
data and does not fully understand its characteristics 
and background.
To enhance DINOv2's capability of hidden representation 
to geophysical data, we have proposed a series of adaptation 
strategies, including selecting a diverse set of adaptation datasets,
choosing appropriate adapter layers,  
designing various decoding modules, and fine-tuning. 
These efforts have enhanced DINOv2's feature representation capability 
in geophysical data, leading to better completeness of targets, 
finer details, and improved separation from the background, 
as shown in the third column of Fig.~\ref{fig:PCA}.

\begin{figure*}[t]
    \centering
    \includegraphics[width=\textwidth]{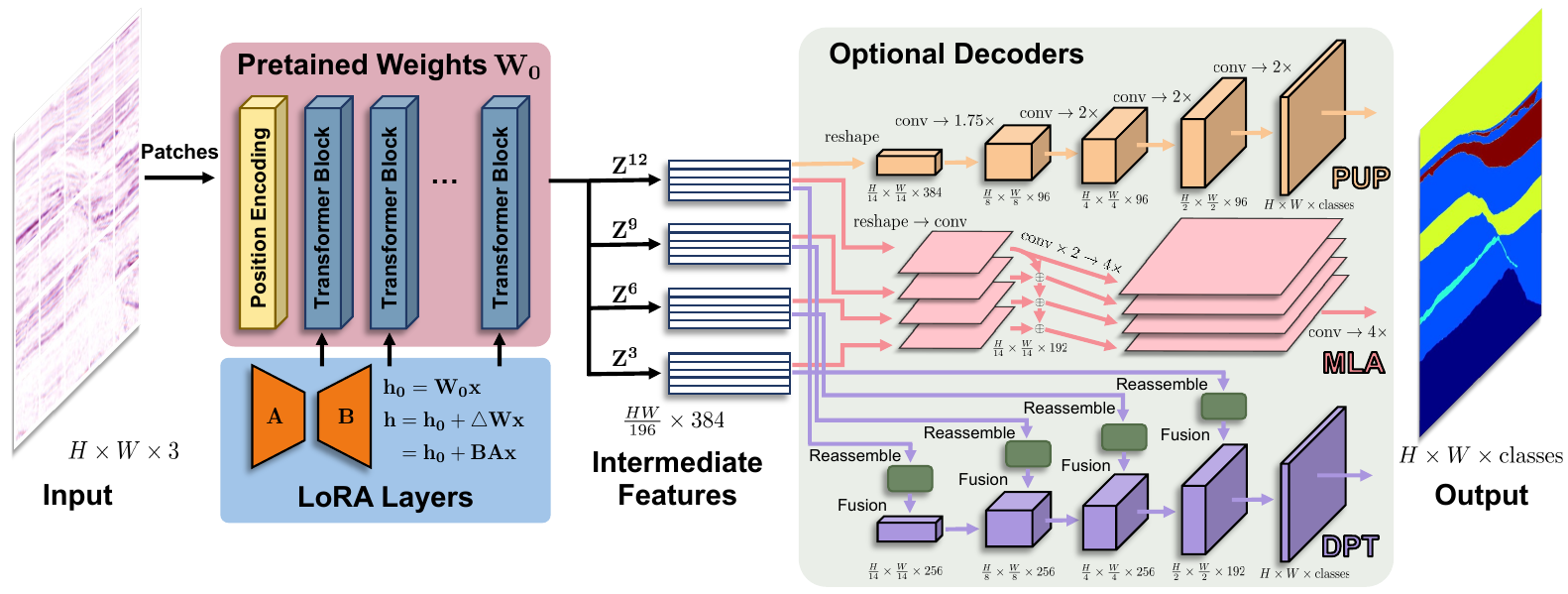}
    \caption{Network architecture of adapting foundation models.
    We designed the adaptation network by feeding the 
    three-channel data into the pre-trained foundation model with a ViT architecture. 
    We employed LoRA layers to efffiently fine-tune the pre-trained ViT and enhance 
    its feature representation of geophysical data. 
    We also explored three different types of decoders 
    (PUP, MLA, and DPT) for mapping the ViT features,
    specifically the features from the 3rd, 6th, 9th, and 12th layers, 
    into the task-specific targets or outputs. 
    This adaptation scheme, involving fine-tuning LoRA layers 
    and custom decoders, enables the development of broad 
    geophysical applications using a pre-trained 
    vision foundation model.}
    \label{fig:decoder}
    \end{figure*}

\subsection{Efficient and generalized adaptation of FM to geophysics }
As shown in Fig.~\ref{fig:workflow}, we designed a general 
workflow for effective cross-domain adaptation of general vision foundation 
models (\hyperref[methods]{Methods}). 
Firstly, to explore how the data types, quantities, 
and sizes affect the adaptation of foundation models, 
we constructed several representative geophysical 
datasets that vary across the three aspects.
They include 
lunar images for crater detection, DAS data 
for seismic event detection, and seismic data for seismic 
facies classification, geobody identification, and deep fault 
detection. Each dataset is tailored to specific geophysical 
tasks and is variant in terms of data types, 
quantities, and sizes (Table~\ref{tab.data} in Appendix). 
Secondly, to adapt the DINOv2 to geophysical data for enhanced 
geophysical feature extraction and representation, 
we effectively and efficiently fine-tuned DINOv2 
using the aforementioned datasets combined with 
the parameter-efficient fine-tuning method, LoRA~\cite{hu2021lora}.
This fine-tuned DINOv2 is used as an powerful encoder to compute 
rich geophysical features as shown in the third column of Fig.~\ref{fig:PCA}.
Thirdly, to translate these features into meaningful outputs 
for each pixel and achieve the task objectives, an appropriate 
decoder is required. 

We explored the impact of utilizing different decoders, 
testing from the simplest linear layer to complex decoders 
like PUP, MLA~\cite{zheng2021rethinking}, and DPT~\cite{ranftl2021vision}. 
The simplest linear layer reflects the encoder's inherent feature 
extraction capabilities, while complex decoders are helpful to 
enhance the downstream performance. The specific decoder configurations
and their connections to the encoder of DINOv2 are 
shown in Fig.~\ref{fig:decoder}.

During the fine-tuning training process,
we employed a weighted Dice loss function which helps improve 
training stability in the presence of class imbalances within 
the geophysical datasets. 
To compare the effectiveness of our method, we used the widely 
referenced Unet~\cite{ronneberger2015u} in the geophysical field as 
the baseline, adopting the same training strategy.
Detailed training parameters for specific tasks are provided in
Table~\ref{tab.training} in Appendix.

\begin{figure*}[htp]
\centering
\includegraphics[width=0.7\textwidth]{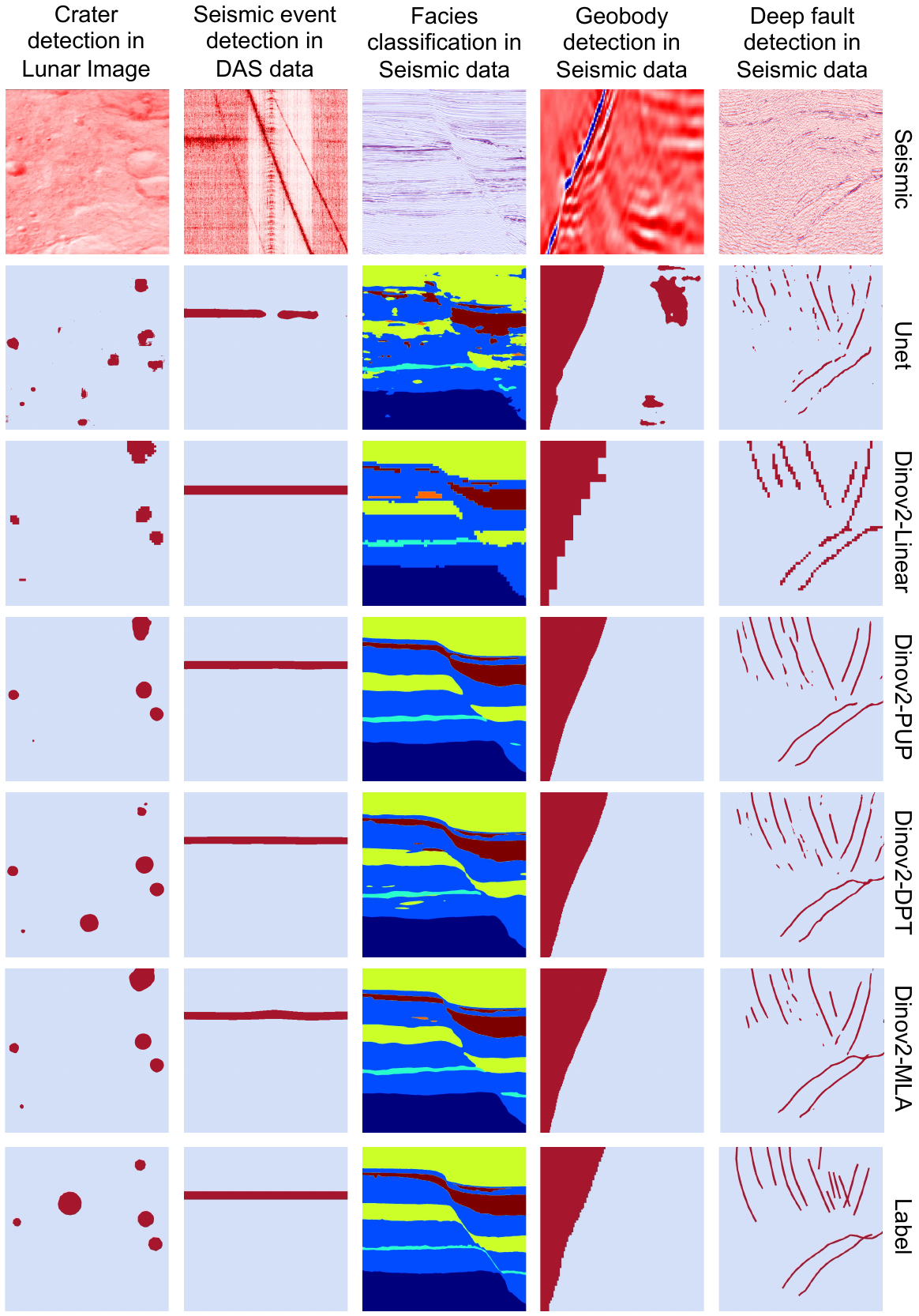}
\caption{Application of the adapted DINOv2 to various geophysical downstream tasks. 
Each column represents a specific downstream task, 
including crater detection in lunar images, seismic event detection 
in DAS data, seismic facies classification, salt dome geobody detection, 
and deep fault detection in seismic data. From top to bottom, 
the rows correspond to the input geophysical data, results from Unet, 
and the adapted DINOv2 encoder with a Linear layer, 
PUP decoder, DPT decoder, and MLA decoder, along with the 
corresponding labels. It is evident that DINOv2, when paired 
with different decoders, achieves good results across all tasks.}
\label{fig:results}
\end{figure*}

\subsection{Performance of adapted FM in geophysical downstream tasks}
For quantitative evaluation of our method during fine-tuning adaptation, 
we used mean Intersection over Union (mIoU) and the mean Pixel Accuracy (mPA), 
both of which are suitable for segmentation tasks. Each task's adaptation 
was conducted on an 80G Nvidia A100 GPU.

\begin{figure*}[htp]
\centering
\includegraphics[width=.5\textwidth]{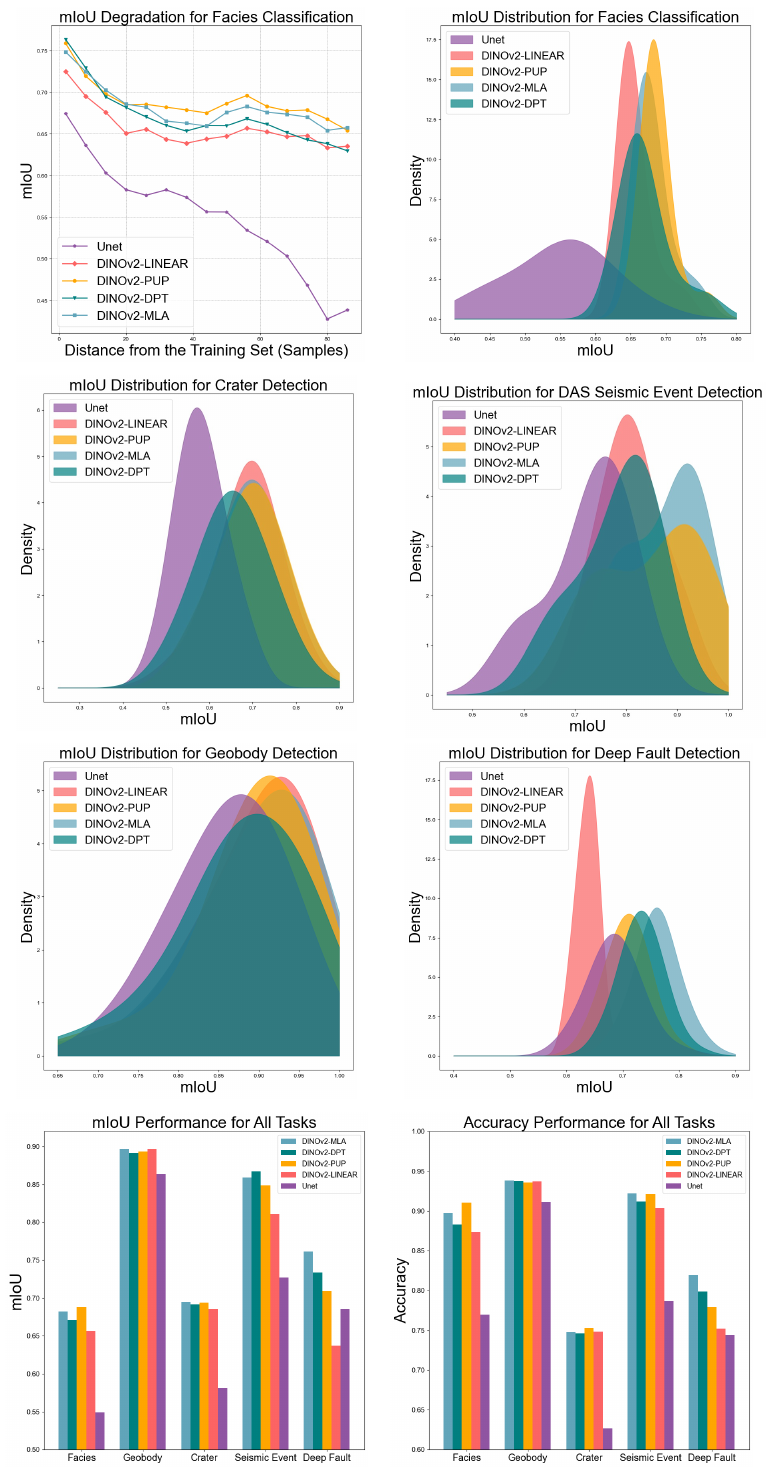}
\caption{Performance metrics on test datasets across all tasks.
For seismic facies classification, 
the mIoU of DINOv2 shows a significantly smaller reduction compared 
to Unet as the distance between the test and training data increases, indicating
that DINOv2 has far superior generalization across diverse data compared to 
Unet. Additionally, we calculated and plotted the mIoU distribution 
for all tasks on the test datasets, further highlighting DINOv2's 
outstanding performance across various tasks. 
Finally, we present the overall mIoU and mPA results, 
showcasing the comprehensive effectiveness of the adapted 
DINOv2 model.}
\label{fig:mIoUAndPA}
\end{figure*}

From the results of adaptation in various geophysical tasks 
(Fig.~\ref{fig:results}), we can see that DINOv2 with any of the four 
decoders performs better than Unet across all five downstream tasks. 
We also displayed the mIoU distribution and mPA results for each task 
sample on the test sets (Fig.~\ref{fig:mIoUAndPA}), the more it is skewed 
to the right and the more concentrated it is, the better the stability 
and performance.
It is worth noting that especially for seismic facies classification, 
the training and test sets were divided into two separate blocks 
from the same 3D data volume (see text S1). 
As the distance 
from the training set increases, the features of the test 
samples differ more from those of the training set. 
In the third column of Fig.~\ref{fig:results} and the first image 
of Fig.~\ref{fig:mIoUAndPA}, we can observe that Unet shows 
a significant performance reduction as the distance increases, 
while our adapted DINOv2 exhibits almost no reduction. This 
indicates that our adapted DINOv2 has much better generalization 
to test data that differ from the training data because it 
learned better feature representations of broad data.
In other tasks, the performance of DINOv2 using the simplest 
linear layer for adaptation is generally comparable to or even exceeds 
that of Unet. This demonstrates that DINOv2 can effectively extract 
and represent features from geophysical data without relying on complex
decoders, as evidenced by the third column of Fig.~\ref{fig:PCA}.

In our experiments, we found that the performance differences 
among the various decoder architectures we used were minimal, 
indicating that the features extracted by DINOv2 are already robust 
and clear enough. Among these decoders, PUP has the fewest parameters 
(0.92M, Table~\ref{tab.params})
and performs consistently well across tasks. Since it recovers directly 
from the last layer of the encoder, it maintains better overall integrity. 
Consequently, we observe that it excels in tasks involving larger targets, 
such as seismic facies classification, crater detection, and seismic geobody detection
(the 2nd, 3rd, and 5th images in Fig.~\ref{fig:mIoUAndPA}).
The MLA decoder introduces multi-scale information from the encoder, 
thus providing better detail recovery compared to PUP. Its overall metrics 
are also high, particularly excelling in seismic event detection 
(0.9222 mPA, Table~\ref{tab.metrics}) and 
deep fault detection (0.8195 mPA, Table~\ref{tab.metrics}). However, 
it has a large number of parameters (10.97M, Table~\ref{tab.params}), 
so computational cost needs to be considered when using it.
As for DPT, it has the most parameters 
(13.58M, Table~\ref{tab.params}) 
and employs 
many multi-scale integration modules, which aids in recovering 
extremely fine details. For instance, in DAS seismic event detection 
(the 5th image in the second column of Fig.~\ref{fig:results}), 
it achieved the highest mIoU (0.8672, Table~\ref{tab.params}). 
However, it introduced some minor noise in seismic facies classification
(the 5th image in the third column of Fig.~\ref{fig:results}). 
The DPT is more suitable for dense prediction than the segmentation tasks in this paper.
More detailed results for each task can be seen in Fig.~\ref{fig:crater}-\ref{fig:fault} in Appendix.
In general, our adapted DINOv2 outperforms Unet across 
various types of geophysical data and different decoder modules
(the last row in Fig.~\ref{fig:mIoUAndPA}). 
This demonstrates the effectiveness of our adaptation, 
and the various tests provide readers with a reference for adaptation. 

\section{Dissusion}\label{sec4}
To overcome the current challenges 
of constructing geophysical foundation models, 
such as the lack of datasets and computational resources, 
by exploring the cross-domain adaptation of computer vision models 
to the geophysical field. We reviewed several mature 
foundation models and selected the most suitable one, 
DINOv2, for geophysical data analysis. 
We explored the capability of DINOv2 (pre-trained on natural images) 
for feature extraction and representation 
of unseen geophysical data. Consequently, we designed a 
comprehensive cross-domain foundation model adaptation workflow, 
conducting detailed experiments and discussions on 
adaptation datasets, fine-tuning methods, and decoding modules. 

With regard to the adaptation datasets, we collected typical geophysical 
datasets, which varies in data features, quantities, and sizes. 
For some small-sample data, such as DAS data with only 115 training 
samples, the adapted FMs can still achieve a high accuracy score 
(0.9222, Table~\ref{tab.metrics}), while the Unet 
with significantly fewer parameters cannot be well-trained from scratch.
This indicates that our adaptation method can achieve the 
adaptation of large models with only a small amount of data.
For datasets of different data sizes, we found that smaller sizes 
yield higher performance metrics 
compared to larger. For example, seismic geobody detection data 
(224$\times$224) and DAS data (512$\times$512), which are smaller in 
size, achieve accuracy metrics above 0.9, 
whereas larger data sizes like crater detection (1022$\times$1022) and 
deep fault detection (896$\times$896) only achieve accuracies 
between 0.7 and 0.8 (Table~\ref{tab.metrics}).
This is because DINOv2's pre-training data resolution is \(448\times448\). 
Consequently, the pre-trained ViT experiences performance 
degradation when inferring data at scalable resolutions~\cite{tian2023resformer}, 
and DINOv2's 
use of absolute position encoding cannot effectively adapt to changes 
in resolution~\cite{fan2024vitar}, 
leading to decreased performance on larger adaptation data sizes 
compared to smaller ones.
Additionally, larger data sizes significantly increase adaptation time 
(adapting 1000 training samples of 1022$\times$1022 crater detection 
takes 23.4 hours), 
which is much longer than the time required for a larger quantity 
of smaller-sized data (adapting 3000 training samples of 224$\times$224 
seismic geobody detection takes only 2.07 hours) 
(Table~\ref{tab.time}).
Overall, the adaptation metrics indicate that our model 
performs well (compared to the Unet), 
demonstrating that our method is applicable to different types of data, 
and additionally, compared to spending several months training a 
complete foundation model, the time cost for our adaptation 
is very low (the fastest takes only 0.22 hours). We hope the performance 
and time consumption on different 
types of datasets with varying sizes and quantities can provide 
readers with a systematic reference.

Most of our best fine-tuning results were achieved using 
full fine-tuning method (Table~\ref{tab.training}), 
as we employed a pre-trained ViT-S/14. 
Fully fine-tuning small models with a low number of samples 
is relatively straightforward and requires a very low 
learning rate (e.g., 1e-5). 
However, we found that in deep fault detection, MLA and DPT 
achieved the best results using LoRA (Table~\ref{tab.training}). 
This is because the boundary features of faults differ 
significantly from the target features in natural images, 
leading to potential collapse with full fine-tuning. 
LoRA allows for stable fine-tuning on few-shot learning, 
which becomes even more evident when using larger encoders.

As for the decoding modules, we conducted tests ranging from the 
simplest linear layer to complex ones such as PUP, MLA, 
and DPT. The outstanding performance of DINOv2 on the 
linear layer indicates that our adaptation has enhanced 
its ability to extract and represent features of geophysical data, 
demonstrating the effectiveness of our adaptation workflow.
To achieve better performance on downstream tasks, 
we can draw insights from the results of the three complex 
decoders. Our experiments showed that for simple 
segmentation tasks such as crater detection, geobody identification, 
and DAS seismic event detection, PUP achieved excellent results. 
It also provided the best continuity in seismic facies classification, 
with fewer parameters (0.92M), making it highly practical. For those 
seeking the best overall performance, MLA is a good choice as 
it recovers details better than PUP. DPT excels 
in tasks that require emphasizing fine details, 
however, its characteristic for dense prediction doesn't 
offer any advantages over MLA in segmentation tasks.
It is necessary to select the appropriate decoder 
based on the specific task type.

Through our experiments, we have gained new insights 
into developing geophysical foundation models and their 
applications. It is well-known that constructing a 
foundation model from scratch requires an extensive dataset 
with rich and representative features for pre-training, 
which is computationally expensive. However, our experiments 
demonstrate that fine-tuning and adapting pre-trained 
foundation models from other domains can also achieve excellent 
results in various geophysical scenarios and tasks. This approach 
requires only a small dataset and is much less computationally 
intensive. Therefore, developing foundation model applications 
in geophysics may not necessarily require building a geophysics-specific 
foundation model from scratch. Fine-tuning and adapting foundation 
models from other domains provide a more efficient and cost-effective 
alternative.

This study has certain limitations. The datasets we selected 
are all related to geophysical segmentation tasks, primarily 
because the visual foundation model DINOv2 was initially 
developed for classification and segmentation of natural images, 
so we naturally chose segmentation tasks. However, geophysics 
encompasses many regression tasks, which are also critical 
areas of research that we did not explore in this paper. 
Additionally, geophysics involves numerous multimodal data 
(e.g., text), which we could integrate into the encoder or 
decoder to fully utilize geophysical data. Furthermore, we 
could incorporate prompt engines like SAM~\cite{kirillov2023segment} 
to enhance the controllability of network inference.

Based on the results and associated analysis, 
we conclude that cross-domain adaptation
of pre-trained foundation models in geophysics
is feasible and advantageous. Our adaptation workflow is 
applicable to various geophysical scenarios and can provide 
valuable insights for future research on foundation models 
in geophysics and other scientific domains.

\section{Methods}\label{methods}
\subsection{Adaptation datasets preparation}
We constructed several representative geophysical 
datasets including lunar images for crater detection, DAS data 
for seismic event detection, and seismic data for seismic 
facies classification, geobody identification, and deep fault 
detection. Each dataset is tailored to specific geophysical 
tasks and is variant in terms of data types, 
quantities, and sizes (Table~\ref{tab.data}). 
The collected datasets exhibit significant feature differences 
and are rich in characteristics, making them representative 
in geophysical segmentation tasks. Craters and geobodies appear as 
block-shaped targets, while seismic events in DAS data, seismic 
facies, and deep faults in seismic data all manifest as 
spatially varying and anisotropic segmentation targets.
These datasets vary in size, ranging from as few as 115 training 
samples to as many as 3000 samples. Additionally, 
the data dimensions are highly varied, spanning from the smallest 
size of $224 \times 224$ to the largest size of $1022 \times 1022$.
The data diversity in types, quantities and sizes 
allows us to comprehensively study the adaptation 
requirements and optimize the adaptation process for the foundation 
models in the geophysics field.

To ensure consistency between these adaptation datasets 
and DINOv2's pre-training data, we converted the single-channel 
geophysical data into three channels. Additionally, due to the 
significant differences in numerical distribution of geophysical data 
across different surveys, we normalized the data to eliminate these 
discrepancies.
During training, we apply a left-right flip 
augmentation to each data sample to increase sample diversity.
This preprocessing step is reversible and does not 
affect practical applications. 
Based on these datasets, we subsequently conducted a series 
of comparative experiments to explore the adaptability 
of the foundation model to different data types, 
varying data quantities, and different data sizes. 
We also compared its performance with that of custom-trained 
deep learning models.

\subsection{LoRA layers for fine-tune}
As mentioned earlier, DINOv2 demonstrates a certain capability 
for feature extraction and representation of geophysical data, 
These capabilities, however, are not yet fully sufficient. 
We therefore use the datasets collected above to fine-tune 
the encoder of DINOv2 for better geophysical feature representation. 
Fully fine-tuning the model would be cost-prohibitive and 
might lead to catastrophic forgetting~\cite{MCCLOSKEY1989109}, so we opted for 
a parameter-efficient fine-tuning (PEFT) approach.
Currently, there are three mainstream PEFT methods. The first is 
Adapter~\cite{houlsby2019parameterefficienttransferlearningnlp}, 
it introduces additional 
layers into the network, and during fine-tuning, only these newly 
added layers are updated. This approach can effectively enhance 
fine-tuning performance but also increases the number of layers in 
the original model.
The second is Prompt tuning~\cite{jia2022visualprompttuning}, this method
involves introducing tokens as prompts into the input or intermediate 
layers. While it can learn the introduced information, the stability 
of the fine-tuning process is relatively poor.
The last is LoRA~\cite{hu2021lora}, it freezes the weights 
of the encoder and incorporates trainable rank decomposition matrices 
into each layer of the transformer, 
significantly reducing the number of training parameters and the time cost.

Considering the training cost and fine-tuning stability, 
we chose LoRA as the fine-tuning method for adapting 
DINOv2 to geophysical data.
LoRA is a commonly 
used PEFT method in the computational field, which adjusts encoder 
parameters in a low-rank setting (the LoRA Layers of Fig.~\ref{fig:decoder}), significantly 
reducing training costs. In the full fine-tuning approach, the number 
of parameters that need to be updated is as large as the initial
network matrix $\mathbf{W_0}$ (assuming $\mathbf{W_0}$ is $ N \times N $). 
Now, the update parameter matrix $\Delta\mathbf{W}$ is decomposed into two 
low-rank matrices: $\mathbf{B}$ ($ N \times r $) and $\mathbf{A}$ ($ r \times N $). 
Since $r$ is typically small (e.g., 8), this significantly reduces the 
number of parameters that need to be updated. As shown in Table~\ref{tab.params} in Appendix, 
using LoRA can reduce the encoder parameters to $1/100$ of the 
original size. 
It is worth noting that while LoRA can stably fine-tune large models, 
full fine-tuning still performs better when it is feasible, as proven 
in previous research~\cite{biderman2024loralearnsforgets}.
Based on comprehensive experiments, we summarized the best fine-tuning methods 
(LoRA and full fine-tuning) for different decoders across various 
tasks (fine-tuning methods of Table~\ref{tab.training} in Appendix).
After fine-tuning, we found that DINOv2 could perform much better 
feature extraction and representation of geophysical data, 
with clearer distinction between targets and background 
and improved target consistency and details, as shown in the third column of Fig.~\ref{fig:PCA}. 
Next, we input the high-dimensional features output by the fine-tuned DINOv2 
into the decoder for downstream tasks.

\subsection{Decoding Module}
To explore the impact of different decoder structures on the 
fine-tuned DINOv2 applied to geophysical downstream tasks, 
we utilize decoders ranging from the simplest linear layer to complex decoders 
like PUP, MLA~\cite{zheng2021rethinking}, and DPT~\cite{ranftl2021vision},
which represent some of the most popular decoding methods 
today, each with its own advantages. PUP performs layer-by-layer 
convolutional upsampling, allowing the encoded features to be 
gradually restored, resulting in more continuous output. This 
can be clearly seen in the fourth row of Fig.~\ref{fig:results}, 
where the seismic facies are continuous and well-distinguished. 
MLA extracts and integrates features encoded at different depths of 
the ViT encoder. As ViT shifts its focus from local to global with incre asing 
network depth~\cite{dosovitskiy2021image}, this decoder can capture 
both global information and local details. Therefore, MLA performs 
better in capturing details of deep faults and seismic facies than PUP 
(the sixth row of Fig.~\ref{fig:results}). However, it requires 
integrating features from multiple layers, significantly 
increasing the network parameters and training cost. Finally, 
DPT, which has a structure similar to Unet~\cite{ronneberger2015u}, 
emphasizes multi-scale detail extraction more than MLA and is 
better suited for dense prediction. It excels at capturing 
fine details of faults but inevitably introduces some minor noise.
In this paper, we compare the performance of these four decoders 
on the adaptation dataset, providing readers with references for 
choosing decoders when working with adaptation datasets.

\subsection{Weighted dice loss}
Due to the imbalance in class sample numbers in 
geophysical segmentation tasks, we adopted the Dice loss function 
with statistical weighting based on class sample numbers. 
The formular is as follows:
\begin{equation}
    \begin{aligned}
            L_{\text{WeightedDice}}&=1-\sum_{k=1}^{C}
            \omega_k
            \cdot \frac{2|P_k \cap G_k|}{|P_k|+|G_k|}, \\
            \omega_k &=  {\frac{\frac{1}{n_k}}{\sum_{k=1}^{C}{\frac{1}{n_{k}}}}}
    \end{aligned}
\end{equation}
where $C$ is the total number of classes in the task, $P_k$ and $G_k$
are the predictions and corresponding labels for the $k-th$ class of the 
current sample, $n_k$ represents the actual number of the $k-th$ class 
in the current sample, and $\omega_k$ denotes the weight of the $k-th$ class
in the current sample. This loss function shows that the larger the number 
of samples in a certain class, the smaller the corresponding weight, 
and vice versa. This approach helps mitigate the issue of class 
imbalance that is common in geophysical datasets.

\subsection{Data availability}
The dataset used in this article have been uploaded 
to Zenodo and are freely available at \url{https://zenodo.org/records/12798750} 
(Guo et al., 2024). 

\subsection{Code availability}
The source codes for adaptation have been uploaded 
to Github and are freely available at 
https://github.com/ProgrammerZXG/Cross-Domain-Foundation-Model-Adaptation.

\section*{Acknowledgements}
This study was strongly supported by the Supercomputing 
Center of the University of Science and Technology of China, 
particularly with the provision of Nvidia 80G A100 GPUs, 
which were crucial for our experiments. 
We also thank SEAM for providing the seismic facies classification dataset, 
TGS for the geobody identification dataset, 
CAS for the crater detection dataset, 
Biondi for the DAS seismic event detection dataset, 
and CIGs for the deep fault detection dataset.

\appendix
\section*{Appendix A: Data Sources}
This paper primarily tests five typical downstream 
segmentation tasks in geophysics, including seismic 
facies classification, geobody identification, 
crater detection, DAS event detection, and deep fault detection. 
The overall overview of the data is shown in Table~\ref{tab.data}, 
and the corresponding mIoU and mPA metrics are presented in Table~\ref{tab.metrics}.

\subsection*{Seismic Facies Classification}
The seismic facies classification dataset is provided by the 
AIcrowd and SEAM-organized competition ``Facies Identification Challenge: 
3-D Image Interpretation by Machine Learning Techniques''~\cite{seam}.
This dataset includes a 3D seismic volume from the publicly available ``Parihaka'' 
seismic survey, annotated by experts and divided into six seismic facies. 
The dimensions of the dataset are \(1006 \times 782 \times 590\). During 
the training process, we split this volume along the last dimension into 
two parts: \(1006 \times 782 \times 500\) and \(1006 \times 782 \times 90\). 
By sampling at intervals of 2, we obtained 250 training samples and 45 
validation samples. This approach effectively eliminates data leakage 
and, due to the significant variability within the data, demonstrates 
the network's generalization capabilities to a considerable extent.

\subsection*{Seismic Geobody (Salt) Identification}

The geobody identification dataset is provided by the Kaggle competition 
``TGS Salt Identification Challenge''~\cite{tgs}, which includes 4,000 seismic data 
samples containing salt domes, along with corresponding labels. Each data 
sample has a size of \(101 \times 101\). We applied bilinear interpolation 
to the seismic data and nearest-neighbor interpolation to the labels, 
resizing them to \(224 \times 224\). From these, 3,000 sample pairs were 
used as the training set, and the remaining 1,000 sample pairs were used 
as the test dataset.

\subsection*{Crater Detection}

The crater data is sourced from the Lunar and Planetary Data Release 
System of the Chinese Academy of Sciences (CAS). We performed 
projections on the data to obtain 1,199 images of the lunar surface, 
each sized \(1022 \times 1022\). Although the data has been 
partially annotated, the corresponding crater labels were sparse 
and incomplete. We manually annotated the data, ultimately 
selecting 1,000 images for the training set and 199 images for 
the test dataset.

\subsection*{DAS Seismic Event Detection}

The DAS dataset is provided by ``An upper-crust lid over the Long Valley magma chamber''~\cite{biondi2023upper}. 
It consists of 143 DAS data samples, each sized \(512 \times 512\). 
We used 115 of these samples as the training set and the 
remaining 28 as the test dataset.

\subsection*{Deep Fault Detection}

The seismic data is derived from several 3D seismic surveys, 
which include numerous deep faults. We annotated these faults 
along the sections and cropped the data to \(896 \times 896\). 
The dataset consists of 1,350 sample pairs, from which we 
selected 1,081 pairs as the training set and 269 pairs as 
the test set.

\newpage

\renewcommand{\thefigure}{S\arabic{figure}}
\setcounter{figure}{0}
\begin{figure*}
    \centering
    \includegraphics[width=\textwidth]{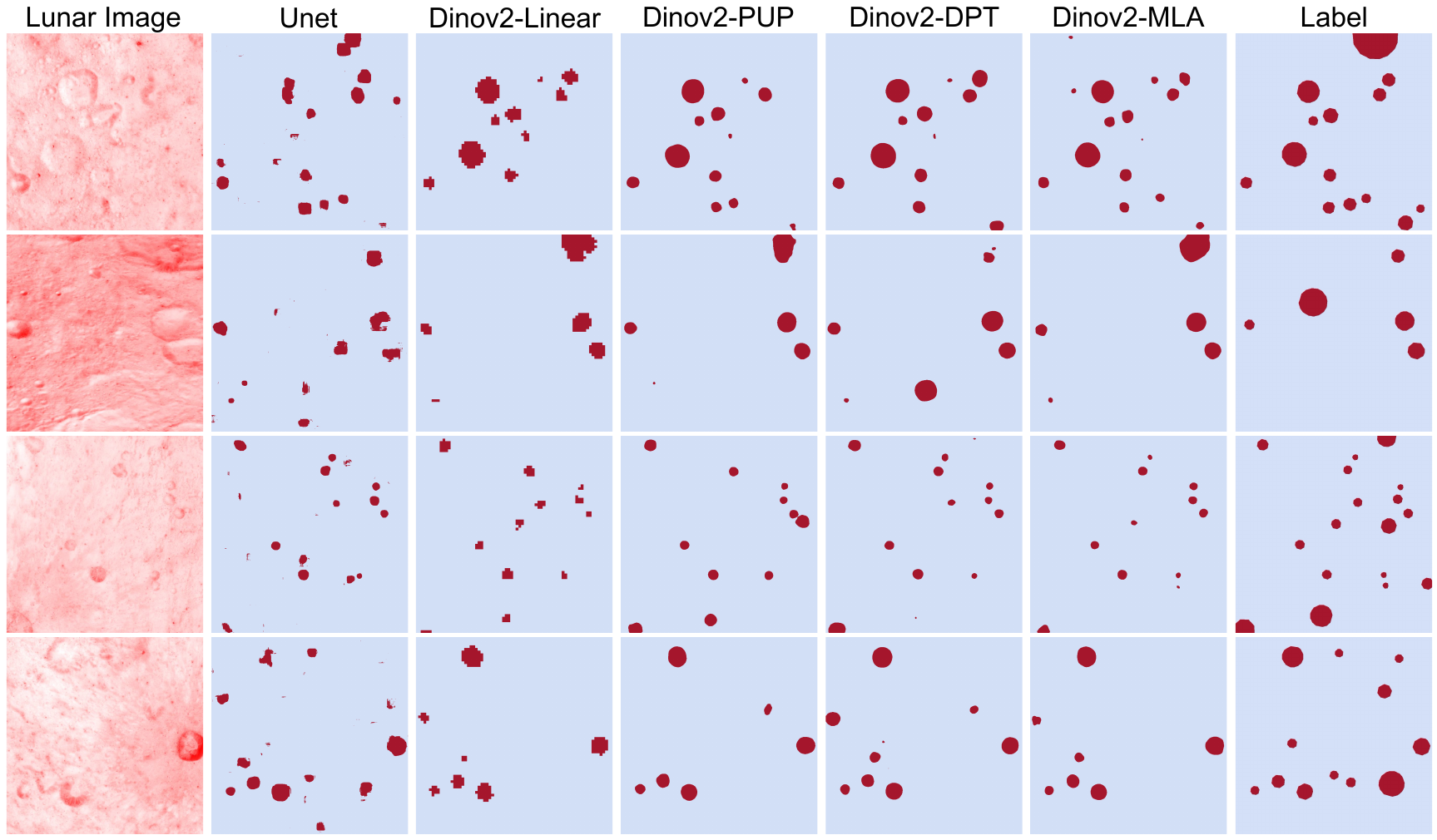}
    \caption{\textbf{More results of various networks in crater detection.}}
    \label{fig:crater}
    \end{figure*}
\clearpage

\begin{figure*}[ht]
    \centering
    \includegraphics[width=\textwidth]{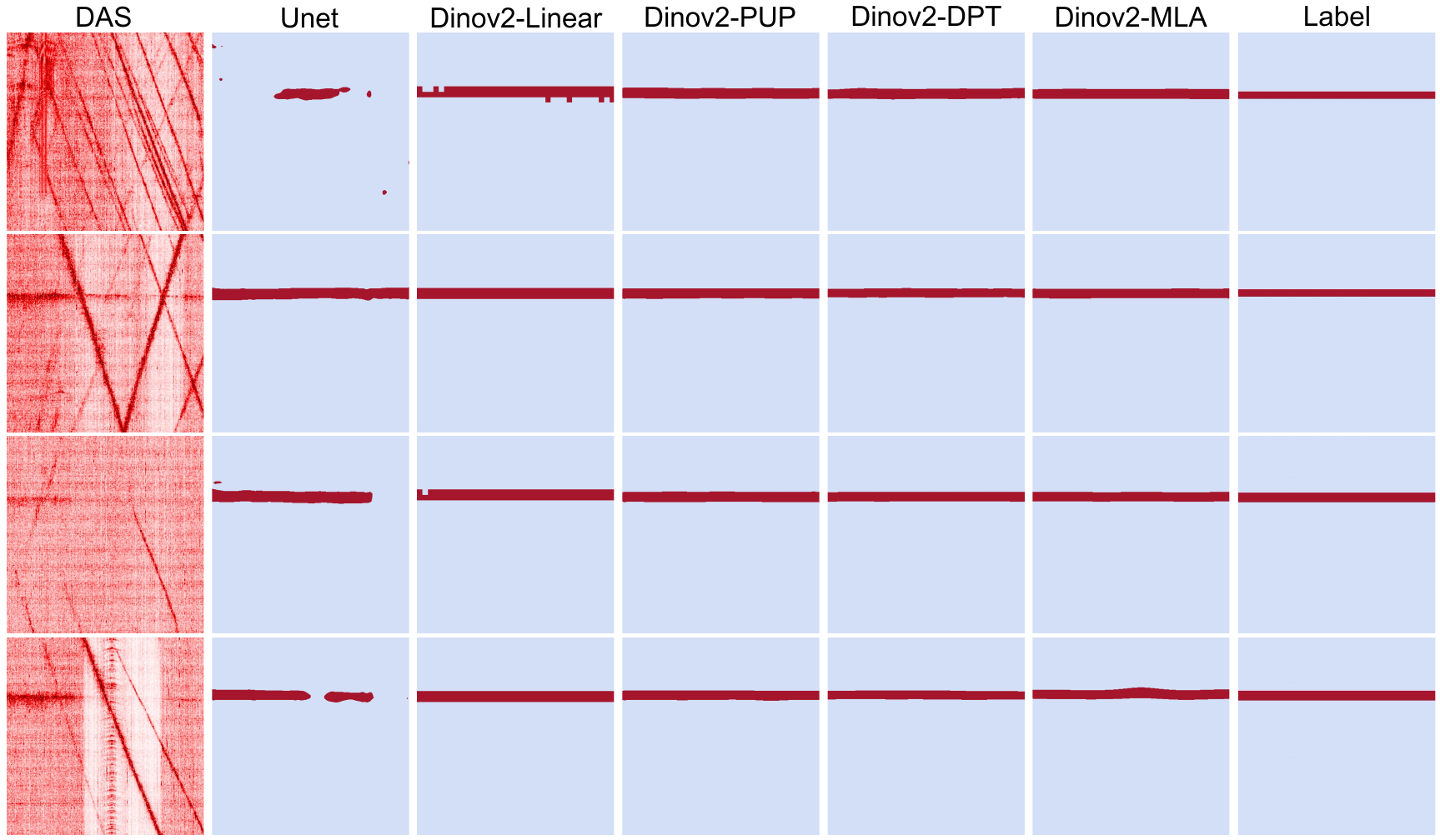}
    \caption{\textbf{More results of various networks in DAS seismic event detection.}}
    \label{fig:das}
    \end{figure*}
\clearpage

\begin{figure*}[ht]
    \centering
    \includegraphics[width=\textwidth]{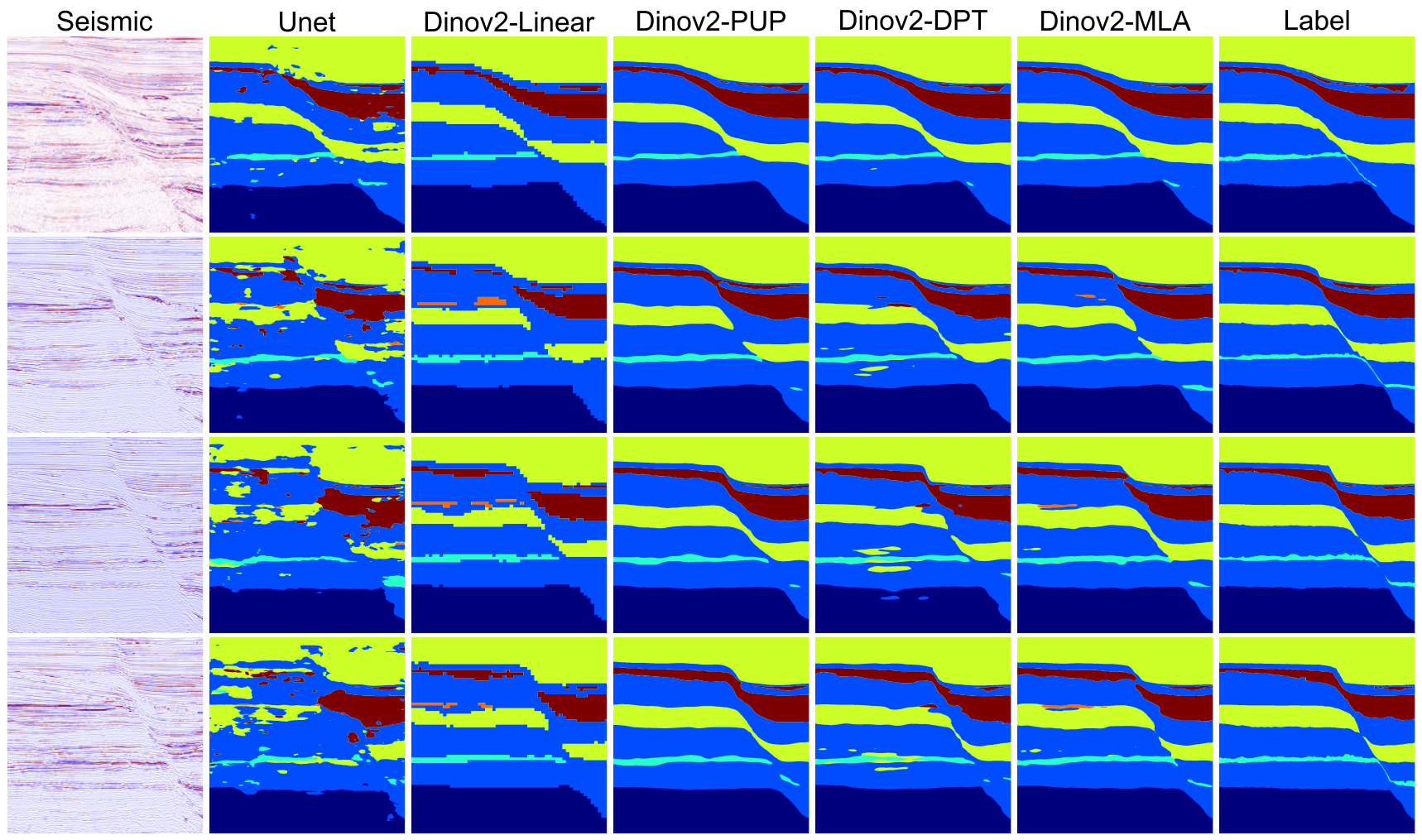}
    \caption{\textbf{More results of various networks in seismic facies classification.}}
    \label{fig:facies}
    \end{figure*}
\clearpage

\begin{figure*}[ht]
    \centering
    \includegraphics[width=\textwidth]{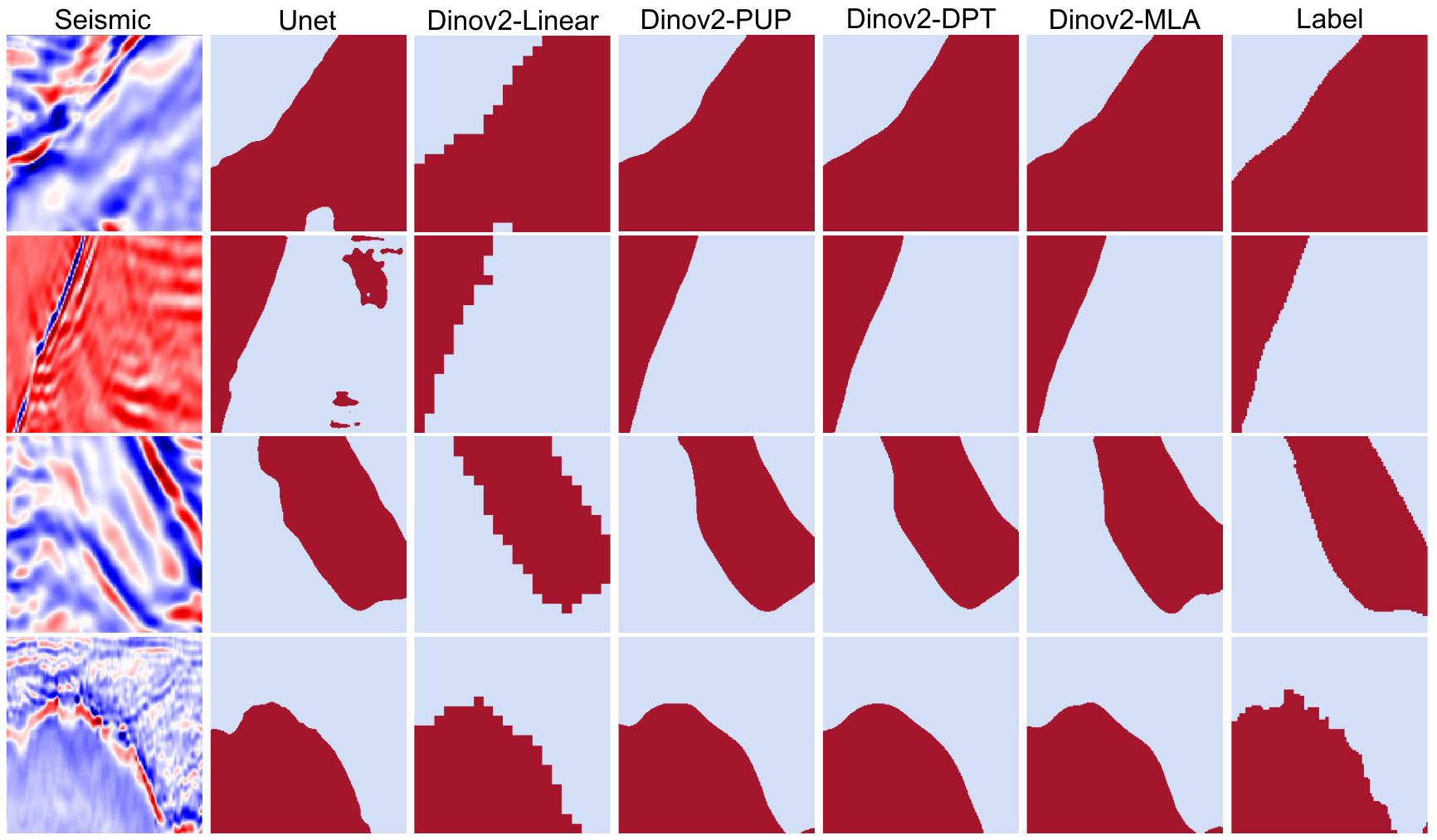}
    \caption{\textbf{More results of various networks in geobody identification.}}
    \label{fig:salt}
    \end{figure*}
\clearpage

\begin{figure*}[ht]
    \centering
    \includegraphics[width=\textwidth]{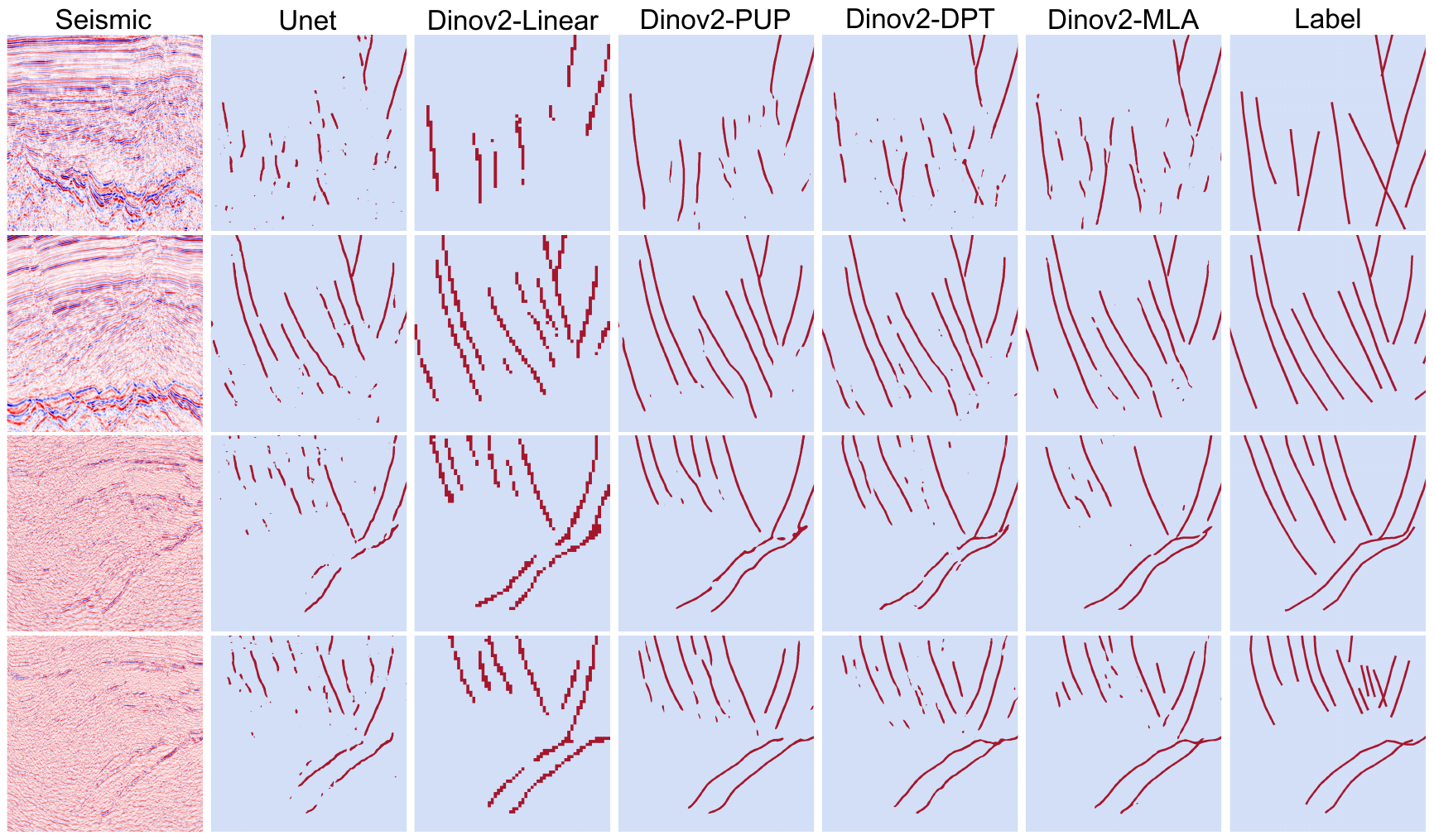}
    \caption{\textbf{More results of various networks in deep fault detection.}}
    \label{fig:fault}
    \end{figure*}
\clearpage

\renewcommand{\thetable}{S\arabic{table}}
\begin{table}[ht]
    \caption{\textbf{Overview of the Datasets}}
    \centering
    \begin{tabular}{lcccc}
    \hline
    Task  & Data Sources & Data Size & \makecell{Training\\ Number} & \makecell{Test\\ Number} \\
    \hline
    \makecell[l]{Seismic Facies\\ Classification}  & \makecell{provided by~\cite{seam}}& $1006\times782$&250 & 45\\
    \hline
    \makecell[l]{Salt Body\\ Identification}  & \makecell{provided by~\cite{tgs}}& $224\times224$&3000&1000\\
    \hline
    \makecell[l]{Crater\\ Detection}  & \makecell{original data provided by\\ CAS, labelled by authors}& $1022\times1022$&1000&199\\
    \hline
    \makecell[l]{DAS Seismic\\ Event Detection} & \makecell{provided by~\cite{biondi2023upper}}& $512\times512$&115&28 \\
    \hline
    \makecell[l]{Deep Fault\\ Detection}    & \makecell{original data provided\\from field surveys,\\ labelled by authors}&  $896\times896$&1081&269\\
    \hline
    \end{tabular}
    \label{tab.data}
\end{table}
\clearpage

\begin{table}[ht]
    \caption{\textbf{Quantitative Metrics for Downstream Tasks}}
    \centering
    \begin{tabular}{lccc}
    \hline
    \multicolumn{4}{c}{\textbf{Mean Intersection over Union(mIoU)}} \\
    Network & \makecell{Seismic Facies \\ Classification} & \makecell{Seismic Geobody \\Identification}
            &\makecell{Crater \\Detection}\\         
    \hline
    Unet          & 0.5490 & 0.8636 & 0.5812\\
    \hline
    DINOv2-LINEAR & 0.6565 & 0.8965 & 0.6857\\
    DINOv2-PUP    & \textbf{0.6885} & 0.8935 & 0.6937\\
    DINOv2-DPT    & 0.6709 & 0.8912 & 0.6917 \\
    DINOv2-MLA    & 0.6826 & \textbf{0.8969} & \textbf{0.6949}\\
    \hline
    Network &\makecell{DAS Seismic\\Event detection}
            &\makecell{Deep Fault \\Detection}\\
    \hline
    Unet          & 0.7271 & 0.6858\\
    \hline
    DINOv2-LINEAR & 0.8112 &0.6372\\
    DINOv2-PUP    & 0.8487 &0.7088\\
    DINOv2-DPT    & \textbf{0.8672} &0.7334\\
    DINOv2-MLA    & 0.8591 &\textbf{0.7613}\\    
    \hline     
    \hline
    \multicolumn{4}{c}{\textbf{Mean Pixel Accuracy(mPA)}} \\
    Network & \makecell{Seismic Facies \\ Classification} & \makecell{Seismic Geobody \\Identification}
            &\makecell{Crater \\Detection}\\     
    \hline
    Unet          & 0.7693 & 0.67 & 0.6265\\
    \hline
    DINOv2-LINEAR & 0.8732 & 0.9374 & 0.7481\\
    DINOv2-PUP    & \textbf{0.9102} & 0.9357 & 0.7529\\
    DINOv2-DPT    & 0.8826 & 0.9377 & 0.7462\\
    DINOv2-MLA    & 0.8975 & \textbf{0.9383} & \textbf{0.7476}\\            
    \hline
    Network &\makecell{DAS Seismic\\Event detection}
    &\makecell{Deep Fault \\Detection}\\
    \hline
    Unet          & 0.7865 & 0.7439\\
    \hline
    DINOv2-LINEAR & 0.9033 &0.7519\\
    DINOv2-PUP    & 0.9210 &0.7793\\
    DINOv2-DPT    & 0.9119 &0.7985\\
    DINOv2-MLA    & \textbf{0.9222} &\textbf{0.8195}\\   
    \hline
    \end{tabular}
    \label{tab.metrics}
    \end{table}
\clearpage

    \begin{table}[ht]
      \caption{\textbf{Training Time (hours)}}
      \centering
      \begin{tabular}{lccc}
      \hline
      Network & \makecell{Seismic Facies \\ Classification} & \makecell{Seismic Geobody \\Identification}
              &\makecell{Crater \\Detection}\\           
      \hline
      Unet          & 0.72 & 0.67 & 4.73\\
      \hline
      DINOv2-LINEAR & 4.62 & 3.22 & 21.20\\
      DINOv2-PUP    & 3.42 & 1.40 & 22.18\\
      DINOv2-DPT    & 1.80 & 2.07 & 23.40\\
      DINOv2-MLA    & 3.55 & 1.77 & 22.24\\        
      \hline
      Network &\makecell{DAS Seismic\\Event detection}
              &\makecell{Deep Fault \\Detection}\\
      \hline
      Unet          & 0.11 & 6.18\\
      \hline
      DINOv2-LINEAR & 0.22 &9.51\\
      DINOv2-PUP    & 0.26 &10.16\\
      DINOv2-DPT    & 0.30 &11.33\\
      DINOv2-MLA    & 0.27 &10.38\\    
      \hline
      \end{tabular}
      \label{tab.time}
      \end{table}
\clearpage

\begin{table}[ht]
    \caption{\textbf{Training Details for Decoder Transfer in Downstream Tasks}}
    \centering
    \begin{tabular}{lccc}
    \hline
    \multicolumn{4}{c}{\textbf{Fine-tuning Methods}}\\
    Network & \makecell{Seismic Facies \\ Classification} & \makecell{Seismic Geobody \\Identification}
            &\makecell{Crater \\Detection}\\          
    \hline
    DINOv2-LINEAR&Full&Full&Full\\
    DINOv2-PUP&Full&Full&Full\\
    DINOv2-DPT&LoRA&Full&Full\\
    DINOv2-MLA&Full&Full&Full\\
    \hline
    Network &\makecell{DAS Seismic\\Event detection}
    &\makecell{Deep Fault \\Detection}\\
    \hline
    DINOv2-LINEAR&Full&Full\\
    DINOv2-PUP&Full&Full\\
    DINOv2-DPT&Full&LoRA\\
    DINOv2-MLA&Full&LoRA\\
    \hline
    \multicolumn{4}{c}{\textbf{Training Parameters Setting}}\\
    Task&optimizer&base\_ lr&batch\_ size\\
    \hline
    Seismic Facies Classification & AdamW & 1e-5 & 3\\
    Seismic Geobody Identification & AdamW & 1e-5 & 32\\
    Crater Detection & AdamW & 1e-5 & 3\\
    DAS SeismicEvent detection& AdamW & 1e-5 & 6\\
    Deep Fault Detection   & AdamW & 1e-5 & 6\\
    \hline
    Task&warmup epochs&lr schedule\\
    \hline
    Seismic Facies Classification &10 & cosine\\
    Seismic Geobody Identification &10 & cosine\\
    Crater Detection &10 & cosine\\
    DAS SeismicEvent detection&10 & cosine\\
    Deep Fault Detection   &10 & cosine\\
    \hline
    \multicolumn{2}{l}{``Full'' means adjusting the entire encoder. }
    \end{tabular}
    \label{tab.training}
    \end{table}
\clearpage

\begin{table}[ht]
    \caption{\textbf{Number of Network Parameters}}
    \centering
    \begin{tabular}{lcc}
    \hline
    Method & Architecture & \makecell{Params\\(Encoder (LoRA/Total)-Decoder)}\\         
    \hline
    Scratch&Unet&4.32M\\
    \hline
    \multirow{4}{*}{DINOv2}&ViT-S/14-LINEAR&0.22/21M-770\\
    &ViT-S/14-PUP&0.22M/21M-0.92M\\
    &ViT-S/14-DPT&0.22M/21M-13.58M\\
    &ViT-S/14-MLA&0.22M/21M-10.97M\\
    \hline
    \end{tabular}
    \label{tab.params}
    \end{table}
\clearpage

\bibliographystyle{plain}
\bibliography{dinov2_ref}

\end{document}